\documentclass[twocolumn, switch]{article}
\usepackage{preprint}
\usepackage{algorithm}
\usepackage{algorithmic}
\usepackage{amsmath, amsthm, amssymb, amsfonts}
\usepackage[numbers,square]{natbib}
\usepackage[utf8]{inputenc}
\usepackage[T1]{fontenc}
\usepackage{xcolor}
\usepackage[colorlinks = true,
            linkcolor = purple,
            urlcolor  = blue,
            citecolor = cyan,
            anchorcolor = black]{hyperref}
\usepackage{booktabs}
\usepackage{nicefrac}
\usepackage{microtype}
\usepackage{lineno}
\usepackage{float}
\usepackage{lipsum}
\usepackage{newfloat}
\DeclareFloatingEnvironment[name={Supplementary Figure}]{suppfigure}
\usepackage{sidecap}
\sidecaptionvpos{figure}{c}
\usepackage{titlesec}
\titlespacing\section{0pt}{12pt plus 3pt minus 3pt}{1pt plus 1pt minus 1pt}
\titlespacing\subsection{0pt}{10pt plus 3pt minus 3pt}{1pt plus 1pt minus 1pt}
\titlespacing\subsubsection{0pt}{8pt plus 3pt minus 3pt}{1pt plus 1pt minus 1pt}
\usepackage{graphicx}
\usepackage{multirow}

\title{DM-KG: A Novel Method for Boosting Spatial Cognition of Vision-Language Models in Street View Imagery}
\usepackage{eso-pic}
\usepackage{tikz}
\usepackage{xcolor}
\PassOptionsToPackage{colorlinks=true,linkcolor=gray,urlcolor=gray}{hyperref}

\newcommand{\AddMyWatermarks}{%
  \begin{tikzpicture}[remember picture, overlay]
    \node[rotate=90, color=gray!60, scale=1] at ([xshift=-4.05in,yshift=0in]current page.center) {%
      \href{https://doi.org/}{Publication doi}%
    };
    \node[rotate=90, color=gray!60, scale=1] at ([xshift=3.9in,yshift=0in]current page.center) {%
      \href{https://doi.org/}{Preprint doi}%
    };
    \node[color=gray!90, scale=1] at ([xshift=0in,yshift=-5in]current page.center) {%
      This is the author's accepted manuscript. The final version will appear in XXXX 2025, \copyright\ XXXX 2025.%
    };
  \end{tikzpicture}%
}
\AddToShipoutPictureBG*{\AddMyWatermarks}

\usepackage{titling}
\usepackage{orcidlink}
\usepackage{footmisc}
\newcommand{\Author}[3]{%
  \textbf{#1}\textsuperscript{#2}\ifx\empty#3\empty\else,\ \orcidlink{#3}\fi%
}
\author{
  \Author{Xinyue Xu}{1}{0009-0005-4353-1412} \and
  \Author{Zheng Zhang}{1,2}{} \and
  \Author{Kunyang Ma}{1}{} \and
  \Author{Ge Zhu}{1}{} \and
  \Author{Lianshuai Cao}{1}{} \and
  \Author{Lei Wang}{1}{} \and
  \Author{Zixuan Li}{1}{} \and
  \Author{Yi Cheng}{1}{}
}
\date{%
  \textsuperscript{1}Institute of Surveying and Mapping, Information Engineering University, Zhengzhou 450001, China\\
  \textsuperscript{2}State Key Laboratory of Geographic Information Science and Technology, Institute of Geographic Sciences and Natural Resources Research, Chinese Academy of Sciences, Beijing 100101, China\\[1em]
  \footnotesize \textbf{Corresponding author:} Yi Cheng (\texttt{chxycy@126.com})\\
}

\begin{document}
\twocolumn[
  \begin{@twocolumnfalse}
\maketitle
\thispagestyle{empty}
\begin{abstract}
As vision-language models (VLMs) are increasingly deployed in geospatial question answering and visual scene understanding, improving their spatial cognition capability on street view imagery for complex logical reasoning has emerged as a key research priority. However, existing VLMs frequently suffer from ``spatial semantic hallucinations'' when perceiving object locations, distances, and directions in real-world street view scenes. Furthermore, such errors are often recalcitrant to tracing and calibration, posing a critical bottleneck for their practical deployment in geospatial tasks. To address this pressing challenge, this study proposes DM-KG (Direction-Metric Knowledge Graph), a structurally grounded spatial representation framework for street view imagery. By explicitly extracting directional and metric relationships between entities from a single 2D image, this framework enhances the spatial reasoning accuracy of VLMs through a structured knowledge graph. Specifically, we integrate panoptic segmentation with metric depth estimation to robustly compute entity-level 3D spatial coordinates. Subsequently, we encode the clock azimuths and Euclidean distances of entity pairs into a JSON-formatted knowledge graph, which is injected into the VLM as an explicit geometric prior to guide spatial reasoning. Experimental results on public spatial question-answering (QA) benchmarks demonstrate that DM-KG reduces the mean absolute error (MAE) in distance estimation by 31.1\% and the mean angular error in direction judgment by 65.8\%, while simultaneously maintaining a high QA success rate. By establishing a complete, augmented reasoning pipeline, this research significantly improves the spatial cognitive capabilities of VLMs in street view scenarios, thereby providing a flexible, generalized, and interpretable framework for geographic visual question answering (GeoVQA) in open environments.
\end{abstract}
\vspace{0.35cm}
  \end{@twocolumnfalse}
]

\section{Introduction}
\label{sec:intro}

Recent advances in Vision-Language Models (VLMs) have demonstrated remarkable capabilities in cross-modal semantic understanding and image content description, unlocking new possibilities for natural language interaction and spatial queries in street view scenarios \cite{ref1,ref2,ref3,ref4}. However, existing VLMs still exhibit significant shortcomings in spatial cognition \cite{ref5,ref6,ref7,ref8}, particularly regarding the spatial geometry, relative positioning, and metric relationships of objects in physical environments \cite{ref9,ref10}. They frequently suffer from ``spatial semantic hallucinations,'' making it difficult to directly support visual question-answering (VQA) tasks \cite{ref11} that demand high spatial accuracy. Therefore, improving how VLMs understand spatial relationships in street view imagery not only enhances their structural cognition of geographic scenes but also provides crucial support for downstream tasks such as geographic VQA, visual localization, and intelligent navigation \cite{ref12}.

To mitigate the pervasive ``semantic hallucination'' issue in current VLMs during spatial cognitive tasks, recent studies have primarily pursued two paths: data-driven approaches and architecture enhancement. Representative works like SpatialVLM \cite{ref13} focus on training models using large-scale spatial reasoning QA datasets, whereas models like SpatialRGPT \cite{ref14} emphasize the integration of 3D visual encoders. Both approaches attempt to endow VLMs with metric-level spatial cognitive abilities. However, the former fundamentally relies on implicit statistical correlations learned from massive 2D datasets, lacking the ability to reason about spatial relationships between specific instances. This makes it inadequate for tasks with explicit directional requirements, such as precise spatial queries \cite{ref15}. Conversely, while architecture-enhanced models have improved specific region parsing, their reasoning process \cite{ref16} remains largely opaque (a ``black box'') in complex, occluded, or expansive street view scenes. Deviations in direction judgment and distance estimation are difficult to trace, attribute, or calibrate \cite{ref17}. Consequently, their spatial cognitive outcomes lack necessary interpretability and maintainability, failing to meet the rigorous deployment demands of real-world applications.

To address these challenges, we propose DM-KG, an interpretable, structurally grounded spatial representation framework for single street view image. This framework recovers entity-level 3D relationships using panoptic segmentation and monocular metric depth estimation, providing external spatial knowledge to enhance VLM VQA capabilities. The main contributions of this paper are as follows:

\begin{enumerate}
\item We propose a VLM spatial reasoning enhancement framework based on external knowledge graph priors which differs from existing spatial enhancement methods that rely heavily on linguistic cues and are prone to statistical biases in datasets. By injecting a structured direction-metric relationship graph into the VLM as a prompt, the framework significantly reduces absolute errors in distance estimation and direction judgment while maintaining a high QA success rate.

\item To overcome the limitations of traditional knowledge graphs that lack geometric topological constraints and existing spatial frameworks that fail to extract fine-grained spatial relationships, we develop a joint panoptic segmentation and depth estimation method for DM-KG construction. This method builds a structured spatial knowledge graph that incorporates both semantic attributes and geometric topological constraints, enabling the explicit and interpretable extraction of spatial relationships within a scene.

\item Given the lack of a standardized evaluation benchmark for monocular depth estimation models in street view entity 3D coordinate calculation---an issue that hinders the advancement of VLM spatial cognition, we construct a benchmark tailored for evaluating monocular depth estimation models in the context of street view entity 3D coordinate calculation. We systematically evaluate four mainstream models across three core distance intervals, providing a standardized selection basis for street view spatial cognition tasks.
\end{enumerate}

\section{Related Work}
\label{sec:related}

Recently, several studies \cite{ref18,ref19} have attempted to reconstruct point clouds or Neural Radiance Fields (NeRFs) from multi-view images and integrate them with Large Language Models (LLMs). However, this paradigm relies heavily on multi-view data, rendering it inapplicable to the single street view image scene that is the focus of this study. Its high computational overhead and the modal gap between 3D geometry and natural language further limit its practical application. In contrast, approaches like SpatialVLM \cite{ref13} adopt a fundamentally different technical route. Instead of relying on explicit 3D reconstruction, they use large-scale automated annotation frameworks to generate massive visual QA datasets, training 2D VLMs to implicitly encode spatial understanding within their parameters. Although this unlocks new capabilities in both qualitative and quantitative spatial QA, inherent limitations remain. First, SpatialVLM relies heavily on linguistic object descriptions. But some research \cite{ref20} indicates that VLMs can often solve certain spatial queries using only text cues or embedded world knowledge, bypassing visual inputs entirely. This suggests that VLMs may rely on statistical biases in the dataset rather than acquiring true visual spatial reasoning. Second, SpatialVLM struggles to accurately specify arbitrary regions, making it difficult to handle queries requiring fine-grained spatial judgments between specific instances. This limitation becomes glaringly obvious in real-world street views when users attempt to describe specific locations using natural language.

To achieve fine-grained image understanding, region-level VLMs have evolved from text-based coordinates to feature alignment. Early methods, such as KOSMOS-2 \cite{ref21}, Shikra \cite{ref22}, MiniGPT-v2 \cite{ref23}, CogVLM \cite{ref24}, SPHINX \cite{ref25}, and LLaVA \cite{ref4}, enable region-based image understanding by incorporating bounding boxes as plain text within the input sequence. However, this makes VLMs overly reliant on their language decoders to understand location, and LLMs are inefficient and inaccurate when processing continuous numerical sequences \cite{ref26}. Subsequent approaches, including VisionLLM \cite{ref27}, GPT4RoI \cite{ref28,ref29}, Ferret \cite{ref30,ref31}, and GLaMM \cite{ref32}, introduce Region of Interest (RoI) alignment feature extraction modules. These map region-level visual features directly to the LLM's word embedding space, mitigating the drawbacks of plain text coordinates. Nevertheless, standard rectangular bounding boxes inevitably capture background noise, leading to inaccurate alignment between target regions and text, thereby increasing the complexity of spatial reasoning. More recent models, such as RegionGPT \cite{ref33} and Osprey \cite{ref34}, go a step further by introducing visual spatial cognition modules capable of directly extracting pixel-level features from arbitrarily shaped masks. This flexible input support allows VLMs to focus precisely on target entities.

It is worth noting that while these region-level VLMs excel at general region understanding, they are not designed to perform precise metric spatial reasoning. They can see and describe a region but struggle to accurately calculate the metric distance and relative direction between two regions. SpatialRGPT \cite{ref14} represents a significant step toward addressing this gap. It learns region representations from 3D scene graphs via a data management pipeline and uses a flexible plug-in module to integrate depth information into existing VLM visual encoders. However, its success rate in metric distance QA remains suboptimal when processing complex, occluded, or distant street view scenes. Furthermore, SpatialRGPT still exhibits significant ``black-box'' characteristics; estimation errors are difficult to trace, leaving the spatial cognitive outcomes without necessary interpretability.

\section{Methods}
\label{sec:methods}

Knowledge graphs \cite{ref35}, with their node-and-edge topological structures, are naturally suited to represent entities and their mutual constraints \cite{ref36}. By encoding direction-metric relationships into a structured, JSON-formatted knowledge graph, we transform ambiguous and noisy visual spatial information into discrete, precise symbolic knowledge. By injecting this explicit, high-confidence spatial graph as a prior context into VLM prompts, the model shifts from hallucination-prone visual guessing to logical deduction based on reliable geometric facts. This not only improves QA accuracy but also ensures the VLM's reasoning process is interpretable and its errors traceable.

\begin{figure}[htbp]
  \centering
  \includegraphics[width=0.95\columnwidth]{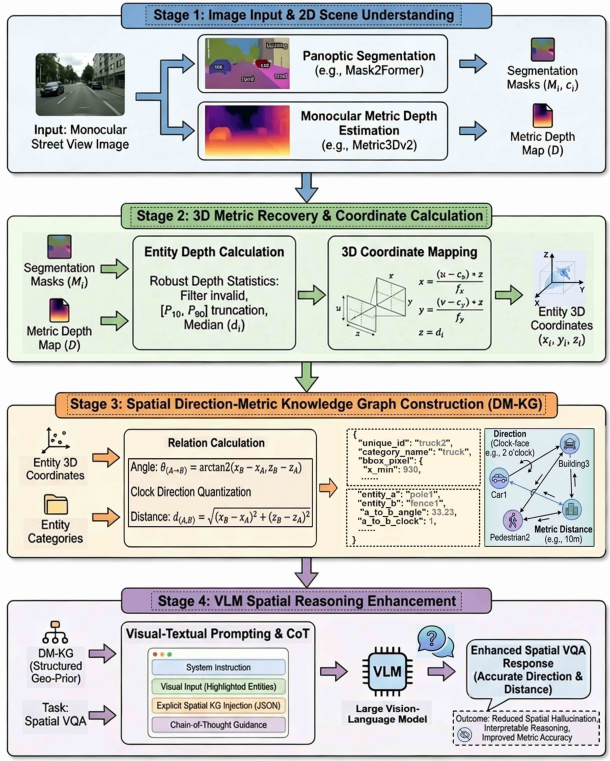}
  \caption{DM-KG Spatial Cognitive Enhancement Framework and Technical Pipeline}
  \label{fig:framework}
\end{figure}

The proposed spatial structured representation framework, DM-KG, is illustrated in Figure~\ref{fig:framework}. It strictly follows four core stages: Image Input and 2D Scene Understanding (Stage~1), 3D Metric Recovery and Coordinate Calculation (Stage~2), Direction-Metric Knowledge Graph Construction (Stage~3), and VLM Spatial Reasoning Enhancement (Stage~4). Each module is detailed below.

\subsection{Entity Semantic Segmentation and Depth Estimation}
\label{subsec:segmentation}

To extract semantically meaningful entities from a single street view image and acquire pixel-level geometric information, this module performs panoptic segmentation and monocular metric depth estimation in parallel.

For entity segmentation, we utilize the Mask2Former panoptic segmentation model. This model simultaneously performs semantic and instance segmentation, assigning category labels to each pixel while distinguishing individual instances within the same category. It outputs a pixel-level panoptic segmentation mask $M_{i}$ and its corresponding category label $c_{i}$. For an input image $I$ of size $H \times W$, the segmentation result is expressed as:
\begin{equation}
\forall p \in I, \text{Panoptic}(p) = \left( s_{i},c_{i} \right)
\label{eq:panoptic}
\end{equation}
where $s_{i}$ is the unique identifier (segment\_id) of the $i$-th entity and $c_{i}$ is its category name (e.g., ``car'', ``vegetation'').

In parallel, a selected monocular depth estimation model (see Section~\ref{sec:experiments} for model evaluation) processes the original image to generate a metric depth map $D \in \mathbb{R}^{H \times W}$, where each pixel value represents the true metric distance (in meters) from that point to the camera's optical center.

\subsection{Calculation of 3D Entity Coordinates}
\label{subsec:coordinates}

After obtaining the segmentation mask and the global depth information for each entity, this stage aims to map the pixel-level features of the entity into physical 3D space.

For entity $i$, the set of depth values within its mask is $\{ D(p) \mid p \in M_{i} \}$. Because depth estimation can produce extreme outliers at object edges, occlusions, or textureless regions, directly averaging the depth is highly prone to error. Therefore, we adopt a robust statistical strategy: first, invalid pixels (depths $\leq 0$ or $> 1000$ m) are filtered out; next, the 10th percentile ($P_{10}$) and the 90th percentile ($P_{90}$) of the valid depths are calculated, and only values within the $[P_{10},P_{90}]$ interval are retained; finally, the median of these remaining values is taken as the representative valid depth $d_{i}$ for the entity.

Given known intrinsic camera parameters, pixel coordinates are converted to 3D coordinates in the camera coordinate system using the pinhole camera model. Assuming camera focal lengths of $f_{x},f_{y}$ (in pixels), principal point coordinates $c_{x},c_{y}$, and a depth value $z = d_{i}$ corresponding to pixel $(u,v)$ in the image, its 3D coordinates $(x,y,z)$ in the camera coordinate system are given by:
\begin{equation}
x = \frac{(u - c_{x}) \cdot z}{f_{x}}, \quad y = \frac{(v - c_{y}) \cdot z}{f_{y}}, \quad z = d_{i}
\label{eq:pinhole}
\end{equation}

For each entity $i$, we take the centroid of its segmentation mask:
\begin{equation}
u_{i} = \frac{1}{|M_{i}|}\sum_{p \in M_{i}} u_{p}, \quad v_{i} = \frac{1}{|M_{i}|}\sum_{p \in M_{i}} v_{p}
\label{eq:centroid}
\end{equation}
as its representative pixel point. Substituting this representative pixel into Equation~\eqref{eq:pinhole} yields the 3D center coordinates $(x_{i},y_{i},z_{i})$. Since our subsequent analysis primarily focuses on spatial constraints on the horizontal plane (X-Z plane), we utilize $(x_{i},z_{i})$ for distance and azimuth calculations.

\subsection{DM-KG Construction}
\label{subsec:dmkg}

To organize the extracted entities and their spatial relationships into a VLM-friendly structured format, we construct the entity-level Direction-Metric Knowledge Graph (DM-KG). Stored in JSON format, this graph explicitly encodes scene geometry via nodes and edges. We define DM-KG as a 2-tuple $\mathcal{G} = \langle \mathcal{E}, \mathcal{R} \rangle$, where $\mathcal{E}$ denotes the set of entities and $\mathcal{R}$ denotes the set of spatial relationships. Let $\mathbb{S}$ be the set of strings, $\mathbb{Z}$ the set of integers, $\mathbb{R}$ the set of real numbers, and $\mathbb{R}^{+}$ the set of positive real numbers:

(1) \textbf{Entity Node Representation:} Each entity $e \in \mathcal{E}$ is defined as a 4-tuple:
\begin{equation}
e = \left\langle id, c, \mathbf{B}_{2D}, \mathbf{P}_{3D} \right\rangle
\label{eq:entity}
\end{equation}
where $id \in \mathbb{S}$ is the unique identifier of the entity; $c \in \mathbb{S}$ is the category name; $\mathbf{B}_{2D} = (x_{\min},y_{\min},x_{\max},y_{\max}) \in \mathbb{Z}^{4}$ represents the 2D bounding box in the image plane; and $\mathbf{P}_{3D} = (x,z) \in \mathbb{R}^{2}$ represents the 3D coordinates in physical space.

(2) \textbf{Relational Edge Representation:} The spatial relationship $r \in \mathcal{R}$ between a source entity $e_{a}$ and a target entity $e_{b}$ is defined as a 5-tuple:
\begin{equation}
r = \left\langle e_{a}, e_{b}, \mathbf{D}_{a \rightarrow b}, \mathbf{D}_{b \rightarrow a}, d \right\rangle
\label{eq:relation}
\end{equation}
where $\mathbf{D}_{a \rightarrow b} = \langle \theta, \tau, s \rangle$ represents the directional attributes from $e_{a}$ to $e_{b}$, including the continuous angle $\theta \in [0,360)$, discrete clock direction $\tau \in \{1,2,\ldots,12\}$, and semantic direction string $s$; $\mathbf{D}_{b \rightarrow a}$ represents the inverse directional attributes from $e_{b}$ to $e_{a}$, structured identically to $\mathbf{D}_{a \rightarrow b}$; and $d \in \mathbb{R}^{+}$ denotes the physical distance between the two entities in meters.

Based on the above structure, the entity nodes of DM-KG comprehensively record the unique identification, category, 2D pixel bounding box, and 3D absolute coordinates. Meanwhile, the relational edges explicitly encode the directions and distances between all entity pairs. To align with human and VLM cognitive habits, our directions utilize the clock direction representation, with the camera's optical center as the origin and the camera's direct front as the 12 o'clock direction. For any two entities $A$ and $B$, the direction angle pointing from $A$ to $B$ is calculated as:
\begin{equation}
\theta_{A \rightarrow B} = \text{arctan2}\left( x_{B} - x_{A}, z_{B} - z_{A} \right)
\label{eq:azimuth}
\end{equation}
This is then quantized to the nearest clock hour (e.g., 15°--45° corresponds to the 1 o'clock direction). The distance is calculated directly using Euclidean distance:
\begin{equation}
d_{A,B} = \sqrt{(x_{B} - x_{A})^{2} + (z_{B} - z_{A})^{2}}
\label{eq:distance}
\end{equation}

By traversing and computing the bidirectional relationships for all entity pairs, the final output JSON graph not only preserves the existential semantics of objects but also thoroughly explicates the 3D physical constraints originally hidden within 2D pixels.

\subsection{Enhancing VLM Spatial Cognition}
\label{subsec:enhancement}

This section is the terminal QA phase of the DM-KG framework. It aims to utilize the DM-KG constructed in the previous steps to calibrate VLM spatial cognitive errors. Instead of altering the underlying VLM architecture, this method transforms implicit spatial guessing into explicit logical reasoning through two mechanisms: visual alignment and structured prompt injection.

When VLMs process complex spatial QA for specific instances, the primary challenge is aligning symbols in the text (e.g., ``car1'') with pixel regions in the image. To solve this, we designed a visual-text bidirectional anchoring mechanism. Rather than feeding the raw image directly, we generate a visually prompted augmented image. The system overlays target entity masks with distinct, semi-transparent identifying colors. On the textual input side, the bounding box coordinates and category attributes of the entity are recorded in the JSON nodes of DM-KG. Through this joint constraint of image-level highlight and text-level bounding box, the VLM can precisely map natural language entities to specific image masks, preventing errors caused by failed entity alignment.

\begin{figure}[htbp]
  \centering
  \includegraphics[width=0.95\columnwidth]{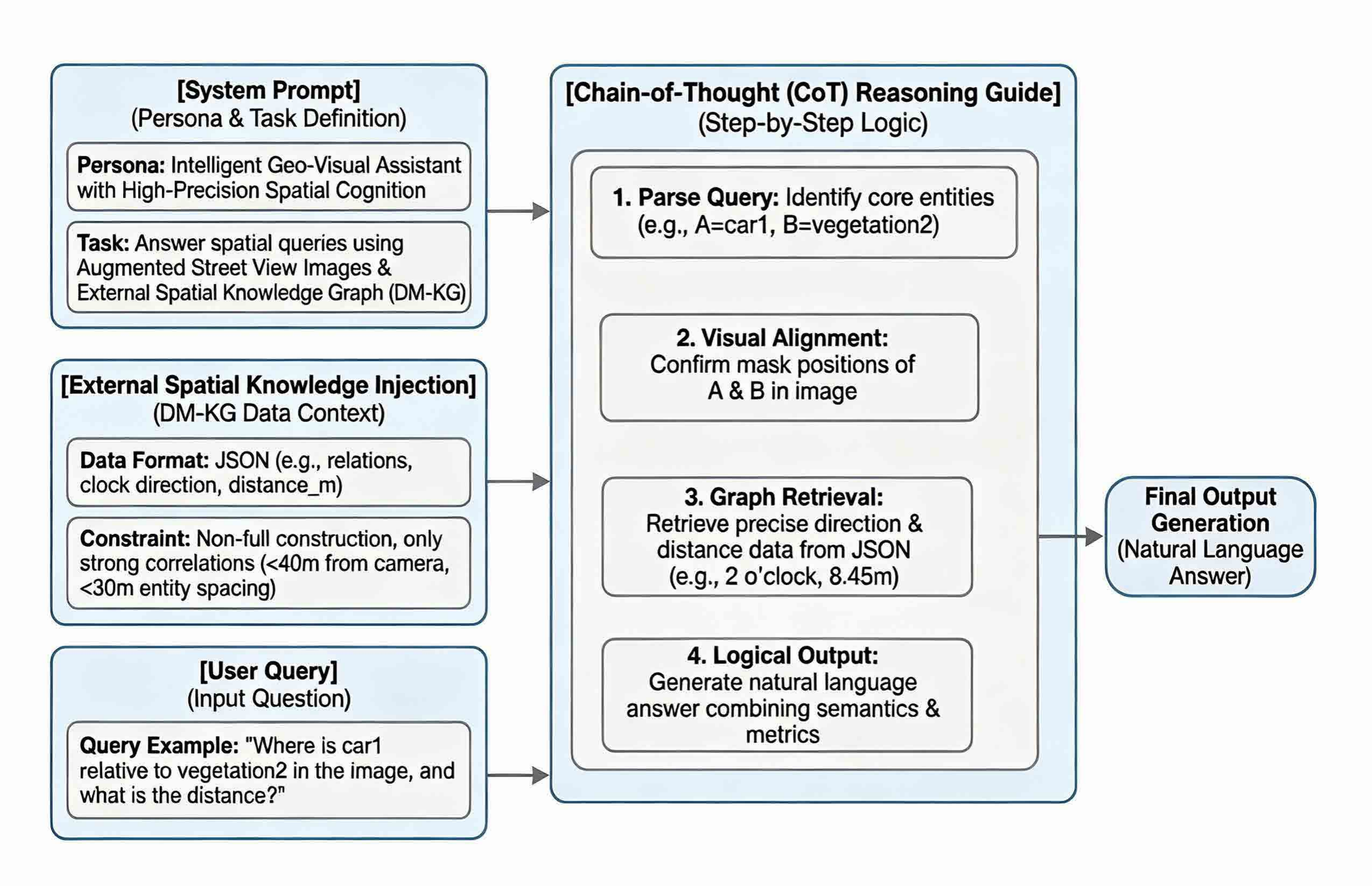}
  \caption{Prompt Template Design}
  \label{fig:prompt}
\end{figure}

To maximize the enhancement effect of DM-KG, we designed a specialized spatial cognition prompt consisting of three modules: system instructions, explicit spatial graph injection, and Chain-of-Thought (CoT) guidance. The prompt template is shown in Figure~\ref{fig:prompt}. In the knowledge graph construction section, to avoid introducing irrelevant noise, the system follows a non-exhaustive construction principle. We only inject relational pairs for entities located within 40 meters of the camera and separated by less than 30 meters.

The strength of this framework lies in dimensionality reduction and decoupling. By injecting DM-KG, we abstract the complex 3D spatial measurement tasks out of the VLM's ``black box'' visual encoder and hand them over to deterministic monocular metric and geometric projection algorithms at the front end. The VLM is then solely responsible for symbolic retrieval and commonsense reasoning tasks. This explicit knowledge injection ensures high QA accuracy while making the geometric rationale behind the answers fully transparent, achieving end-to-end interpretability.

\section{Experiments}
\label{sec:experiments}

\subsection{Experimental Evaluation Metrics Design}
\label{subsec:metrics}

Our depth estimation model evaluation uses 2,000 images from the KITTI \cite{ref37} street view dataset. Given the diversity of street scenes, Google Street View \cite{ref38} and NuScenes \cite{ref39} datasets were included as supplementary validation sources for spatial visual analysis. Quantitative evaluation of spatial reasoning relies on the SpatialRGPT-Bench baseline, which features visual QA samples from KITTI and NuScenes, comprehensively reflecting VLM capabilities in real-world environments.

The evaluated models represent current mainstream depth estimation models: Metric3Dv2(s/g/l) \cite{ref40}, DepthAnythingV2(s/b/l) \cite{ref41}, DepthPro \cite{ref42}, and UniDepthV2(s/b/l) \cite{ref43}. These recently released models have been rigorously tested in academic and industrial settings. Notably, they handle camera focal lengths differently: DepthAnythingV2 lacks an explicit focal length input interface, relying entirely on implicit estimation; Metric3Dv2 requires precise focal length values to generate valid depth maps; the other models support a dual-mode operation, autonomously estimating focal lengths internally or accepting external camera parameters for improved precision.

The accuracy of depth estimation is quantified by Mean Absolute Error (MAE, in meters) and Absolute Relative Error (AbsRel). For the spatial reasoning VQA tasks, the evaluation criteria are set as the success rates of distance estimation and direction judgment under a relative error tolerance of $\leq \pm 25\%$, supplemented by MAE and AbsRel metrics for error analysis.
\begin{equation}
\text{MAE} = \frac{1}{N}\sum_{i=1}^{N}\left| \text{depth}_{\text{pred}}^{(i)} - \text{depth}_{\text{true}}^{(i)} \right|
\label{eq:mae}
\end{equation}
\begin{equation}
\text{AbsRel} = \frac{1}{N}\sum_{i=1}^{N}\frac{\left| \text{depth}_{\text{pred}}^{(i)} - \text{depth}_{\text{true}}^{(i)} \right|}{\text{depth}_{\text{true}}^{(i)}}
\label{eq:absrel}
\end{equation}
where $N$ is the total number of pixels, $\text{depth}_{\text{pred}}^{(i)}$ and $\text{depth}_{\text{true}}^{(i)}$ denote the predicted depth and ground truth depth of the $i$-th point, respectively.

\subsection{Selection of Entity Depth Estimation Models}
\label{subsec:depth}

\begin{figure}[htbp]
  \centering
  \includegraphics[width=0.85\columnwidth]{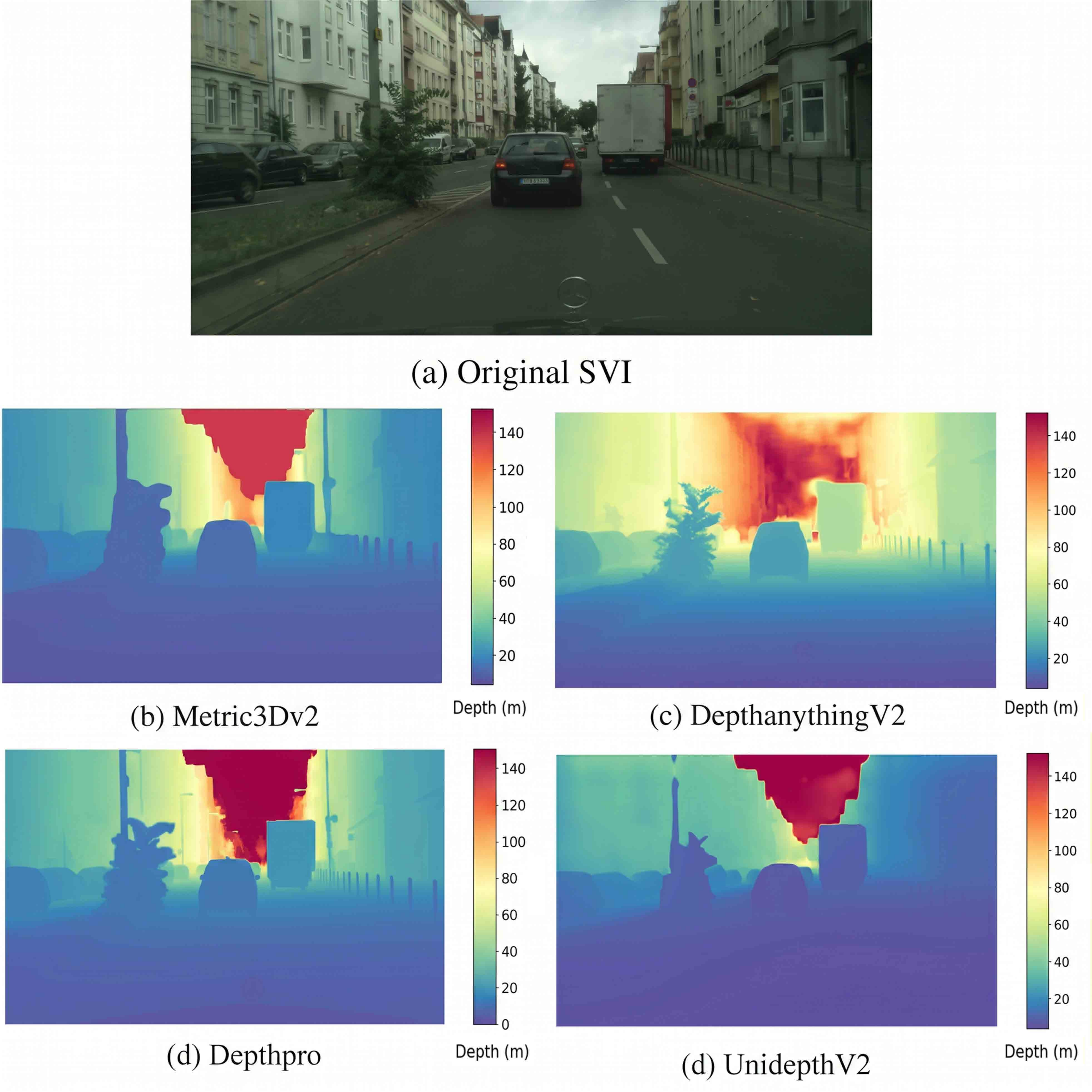}
  \caption{Qualitative comparison of predictive results of representative metric depth estimation models on street view imagery (Unit: meters)}
  \label{fig:depth_qual}
\end{figure}

This experiment evaluates four representative models for 3D spatial coordinate calculations: Metric3Dv2, DepthAnythingV2, DepthPro, and UniDepthV2. Figure~\ref{fig:depth_qual} illustrates the qualitative depth map predictions. DepthAnythingV2 (Figure~\ref{fig:depth_qual}c) struggles with sky regions and distant backgrounds, but its depth restoration for near-camera objects is similar to DepthPro (Figure~\ref{fig:depth_qual}d), both generating reasonable structures. DepthPro produces exceptionally fine contours, sharply resolving grass and vehicle side mirrors. UniDepthV2 (Figure~\ref{fig:depth_qual}e) and Metric3Dv2 (Figure~\ref{fig:depth_qual}b) fail to reconstruct these minute structural features, though they show high consistency in overall depth hierarchy.

We further quantified precision across distance intervals, with and without camera focal length priors. Detailed quantitative results are summarized in Table~\ref{tab:depth}. Since the KITTI dataset lacks valid ground truth within 0--5 meters, our calculations begin at 5 meters. The experimental data reveal a noteworthy phenomenon: without providing focal lengths, base models paradoxically outperform large models. However, when focal lengths are provided, large models universally achieve superior accuracy. We hypothesize that for large models, the advantage in parameter quantity can only be transformed into accuracy advantage after providing focal length input for constraints. Otherwise, the redundant parameters will become noise.

\begin{table*}[t]
  \caption{Comparison of depth estimation accuracy across models with/without explicit focal length input}
  \label{tab:depth}
  \centering
  \small
  \begin{tabular}{llcccccccc}
    \toprule
    \multirow{2}{*}{Model} & \multirow{2}{*}{Focus} & \multicolumn{2}{c}{Total} & \multicolumn{2}{c}{5--10m} & \multicolumn{2}{c}{10--20m} & \multicolumn{2}{c}{$>$20m} \\
    \cmidrule(lr){3-4} \cmidrule(lr){5-6} \cmidrule(lr){7-8} \cmidrule(lr){9-10}
    & Input & MAE(m) & AbsRel & MAE(m) & AbsRel & MAE(m) & AbsRel & MAE(m) & AbsRel \\
    \midrule
    Metric3Dv2-s      & $\surd$ & 1.18 & 0.051 & 0.28 & 0.036 & 0.59 & 0.041 & 2.88 & 0.080 \\
    Metric3Dv2-l      & $\surd$ & 0.88 & 0.042 & 0.30 & 0.038 & \textbf{0.55} & \textbf{0.039} & 1.90 & 0.051 \\
    Metric3Dv2-g      & $\surd$ & \textbf{0.82} & \textbf{0.040} & \textbf{0.26} & \textbf{0.033} & 0.57 & 0.040 & \textbf{1.73} & \textbf{0.046} \\
    DepthAnythingV2-s & $\times$ & 1.78 & 0.080 & 0.36 & 0.042 & 1.30 & 0.089 & 3.95 & 0.112 \\
    DepthAnythingV2-b & $\times$ & 1.46 & 0.064 & \textbf{0.28} & 0.037 & 0.90 & 0.061 & 3.42 & 0.097 \\
    DepthAnythingV2-l & $\times$ & 1.50 & 0.073 & 0.41 & 0.052 & 0.99 & 0.068 & 3.34 & 0.101 \\
    DepthPro          & $\times$ & 1.33 & 0.085 & 0.87 & 0.110 & 1.08 & 0.078 & \textbf{2.70} & \textbf{0.078} \\
                      & $\surd$ & 1.15 & 0.050 & 0.27 & 0.034 & 0.60 & 0.042 & 2.20 & 0.064 \\
    UniDepthV2-s      & $\times$ & 2.09 & 0.101 & 0.53 & 0.066 & 1.50 & 0.104 & 4.51 & 0.134 \\
                      & $\surd$ & 1.39 & 0.101 & 0.52 & 0.067 & 0.87 & 0.061 & 2.97 & 0.083 \\
    UniDepthV2-b      & $\times$ & \textbf{1.22} & \textbf{0.056} & 0.29 & \textbf{0.036} & \textbf{0.79} & \textbf{0.055} & 2.75 & 0.079 \\
                      & $\surd$ & 1.15 & 0.054 & 0.36 & 0.050 & 0.70 & 0.050 & 2.56 & 0.070 \\
    UniDepthV2-l      & $\times$ & 1.68 & 0.085 & 0.54 & 0.069 & 1.23 & 0.086 & 3.48 & 0.101 \\
                      & $\surd$ & 1.00 & 0.050 & 0.38 & 0.050 & 0.63 & 0.045 & 2.12 & 0.057 \\
    \bottomrule
  \end{tabular}
\end{table*}

\begin{figure}[htbp]
  \centering
  \includegraphics[width=0.95\columnwidth]{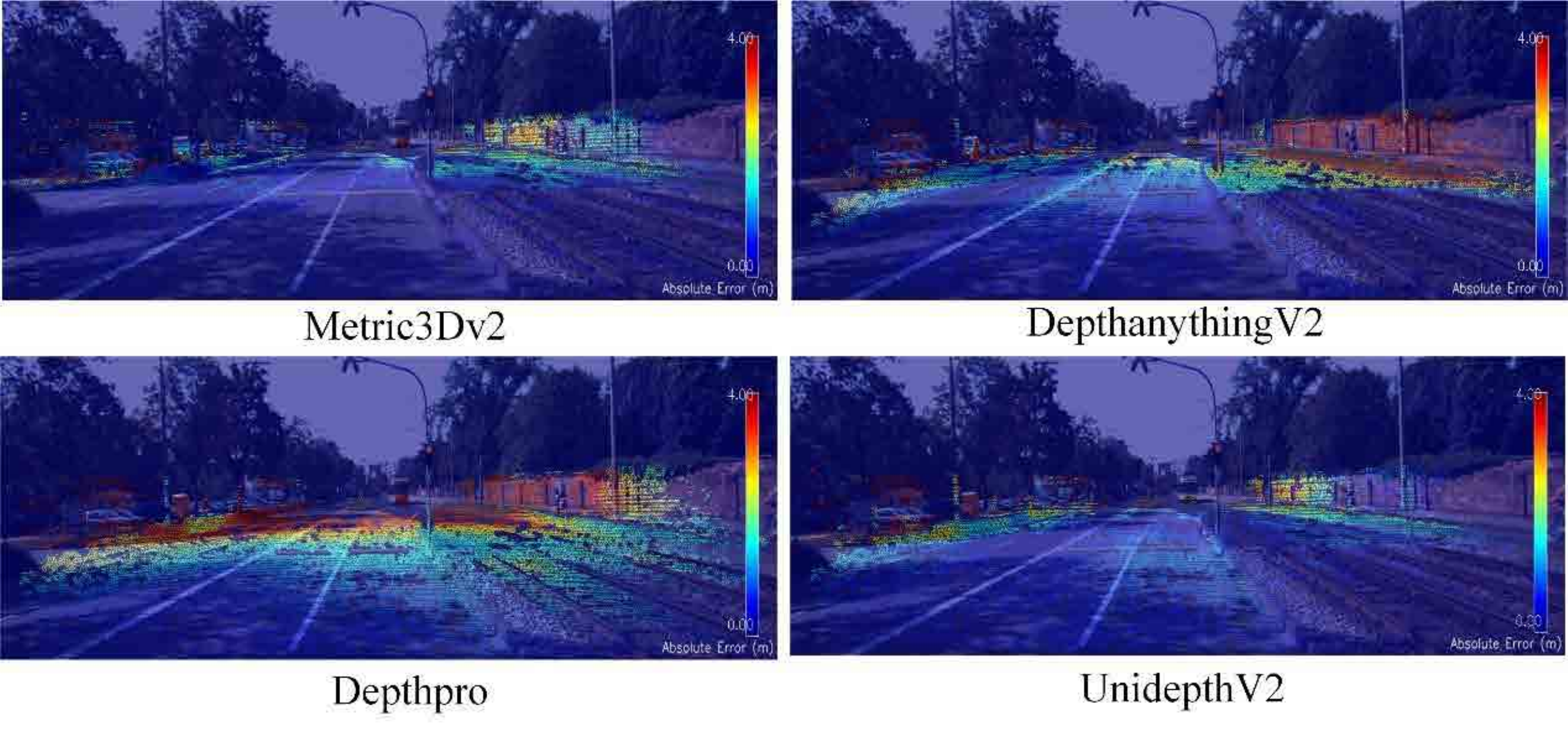}
  \caption{Heatmap visualizations of MAE and AbsRel for representative depth estimation models}
  \label{fig:heatmap}
\end{figure}

\begin{figure}[htbp]
  \centering
  \includegraphics[width=0.95\columnwidth]{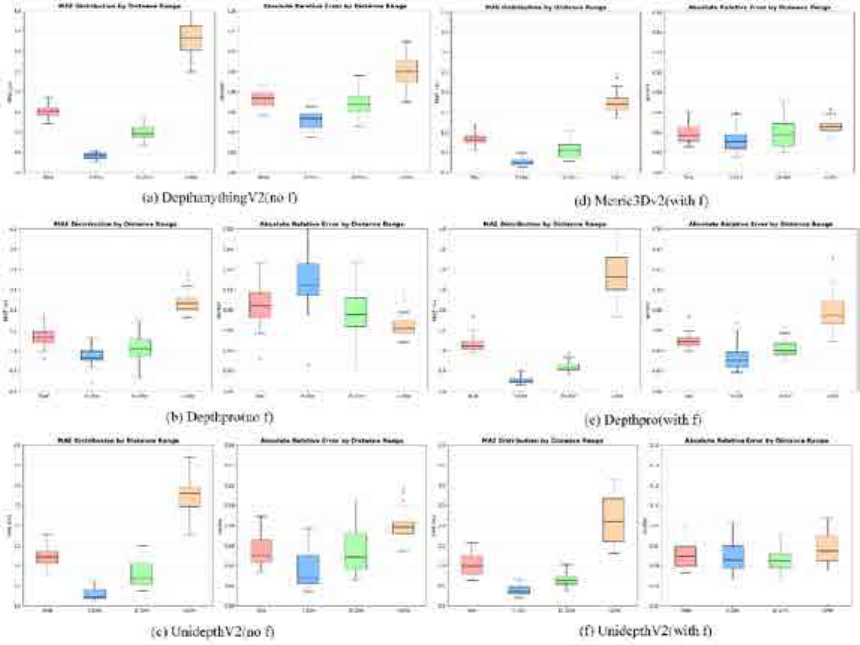}
  \caption{Box plots illustrating MAE and AbsRel error distributions across varying distance intervals with and without camera focal length inputs}
  \label{fig:boxplot}
\end{figure}

\begin{figure}[htbp]
  \centering
  \includegraphics[width=0.95\columnwidth]{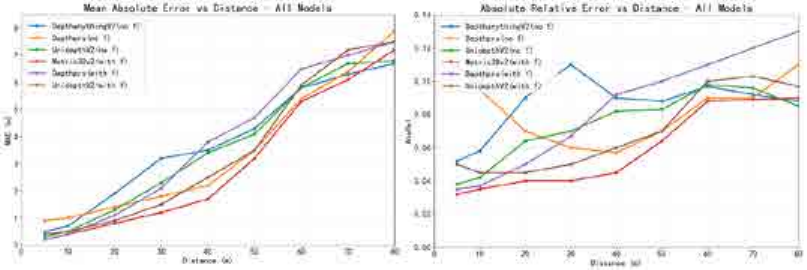}
  \caption{Line chart comparisons demonstrating the impact of focal length priors on depth estimation accuracy across continuous distance intervals}
  \label{fig:line}
\end{figure}

To ensure a fair and intuitive visual comparison, we globally unified the coordinate axis ranges and error heatmap color scales. The upper limits for MAE and AbsRel in the heatmaps (Figure~\ref{fig:heatmap}) and box plots (Figure~\ref{fig:boxplot}) were fixed at 4m and 0.16, respectively, while the line chart axes (Figure~\ref{fig:line}) maxed out at 8m and 0.14. Both the quantitative data and Figures~\ref{fig:heatmap}, \ref{fig:boxplot}, and \ref{fig:line} show that depth estimation errors escalate significantly as target distances increase. Errors in the $>$20m interval are substantially higher than in the 5--20m range, proving that far-distance estimation is the primary bottleneck. Figure~\ref{fig:boxplot} presents box plots of MAE and AbsRel error distributions across distance intervals with and without focal length inputs, indicating that focal length priors effectively compress error dispersion. Line chart comparisons in Figure~\ref{fig:line} further confirm that explicit camera focal lengths as prior inputs significantly enhances the absolute precision, especially manifesting as a sharp decline in the near-to-mid distance range (5--25m).

Selecting depth estimation models for spatial 3D coordinate calculation requires considering absolute metric depth precision, full-distance error consistency, camera intrinsic adaptability, and real-time requirements. For scenarios where precise focal lengths are available, like street view 3D reconstruction or GIS spatial positioning, Metric3Dv2-g is the superior choice. It achieves optimal precision and robustness across all distances, minimizing error propagation during coordinate conversion. Conversely, for scenarios lacking precise camera intrinsics such as uncalibrated legacy imager, UniDepthV2-b is highly recommended. Its built-in focal length prediction module compensates for unknown intrinsics, resulting in minimal performance gaps whether focal lengths are provided or not. If focusing on far-distance targets without camera parameters, DepthPro is ideal due to its superior error control beyond 20 meters. For lightweight, real-time edge computing, Metric3Dv2-s or UniDepthV2-b offer highly cost-effective alternatives, maintaining global MAEs under 1.2m with lower computational overhead.

\subsection{DM-KG Construction Experiments}
\label{subsec:dmkg_exp}

\begin{figure}[htbp]
  \centering
  \includegraphics[width=0.95\columnwidth]{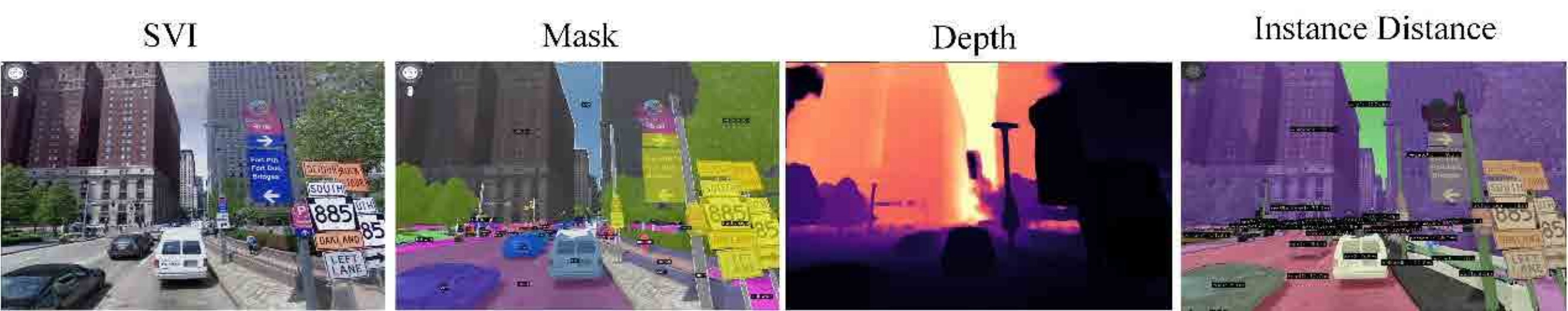}
  \caption{Qualitative visualizations of the joint extraction process integrating panoptic segmentation and metric depth estimation}
  \label{fig:joint}
\end{figure}

\begin{figure}[htbp]
  \centering
  \includegraphics[width=0.85\columnwidth]{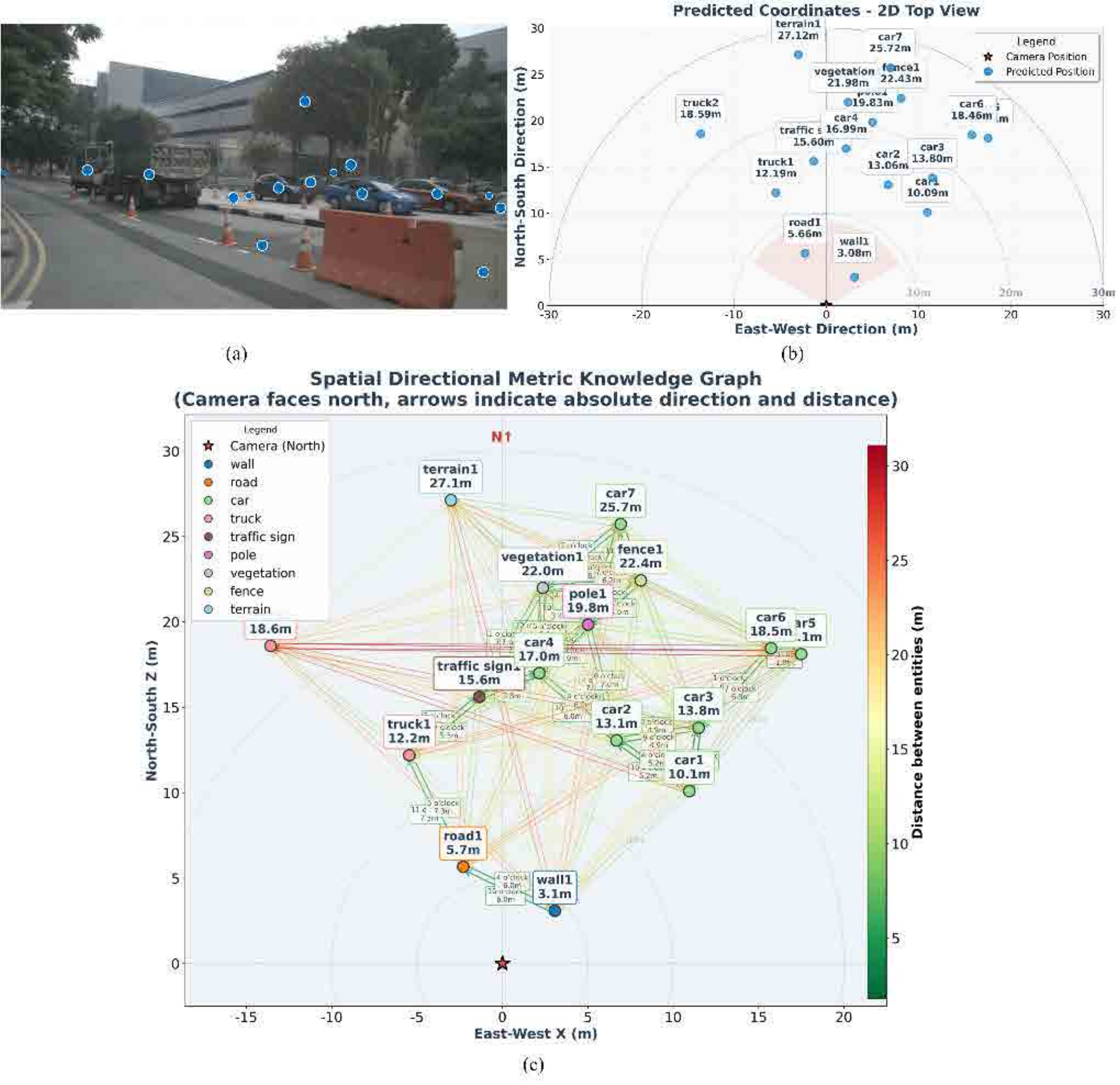}
  \caption{From 2D imagery to 3D representation: comparison of image pixel coordinates and structured DM-KG spatial relationships}
  \label{fig:3dvis}
\end{figure}

For panoptic segmentation, we employed the Mask2Former architecture pre-trained on Cityscapes. As one of the most general benchmarks for street view semantic understanding, Cityscapes' 19 defined target categories exhibit strong prior consistency with urban scene semantics. Figures~\ref{fig:joint} and \ref{fig:3dvis} provide visualizations of the joint extraction tasks involving panoptic segmentation and depth estimation. To ensure high precision, we selected Metric3Dv2, the winner from the previous step, as our depth estimation model. Mask2Former accurately distinguishes semantic entities in street view images, such as road surfaces, vehicles, pedestrian and traffic signs, at both the instance and semantic levels. Subsequently, these masks are then mapped to the depth module's output space to calculate 3D coordinates of each segmented entity. A comparison between the 2D image coordinates (Figure~\ref{fig:3dvis}a,b) and the resulting 3D DM-KG visualizations (Figure~\ref{fig:3dvis}c) demonstrates that the extracted directional and metric relationships closely align with the physical geometry of the scene, validating the effectiveness of our framework.

\subsection{VLM-Based Spatial Reasoning Experiments}
\label{subsec:vlm_exp}

\begin{figure}[htbp]
  \centering
  \includegraphics[width=0.85\columnwidth]{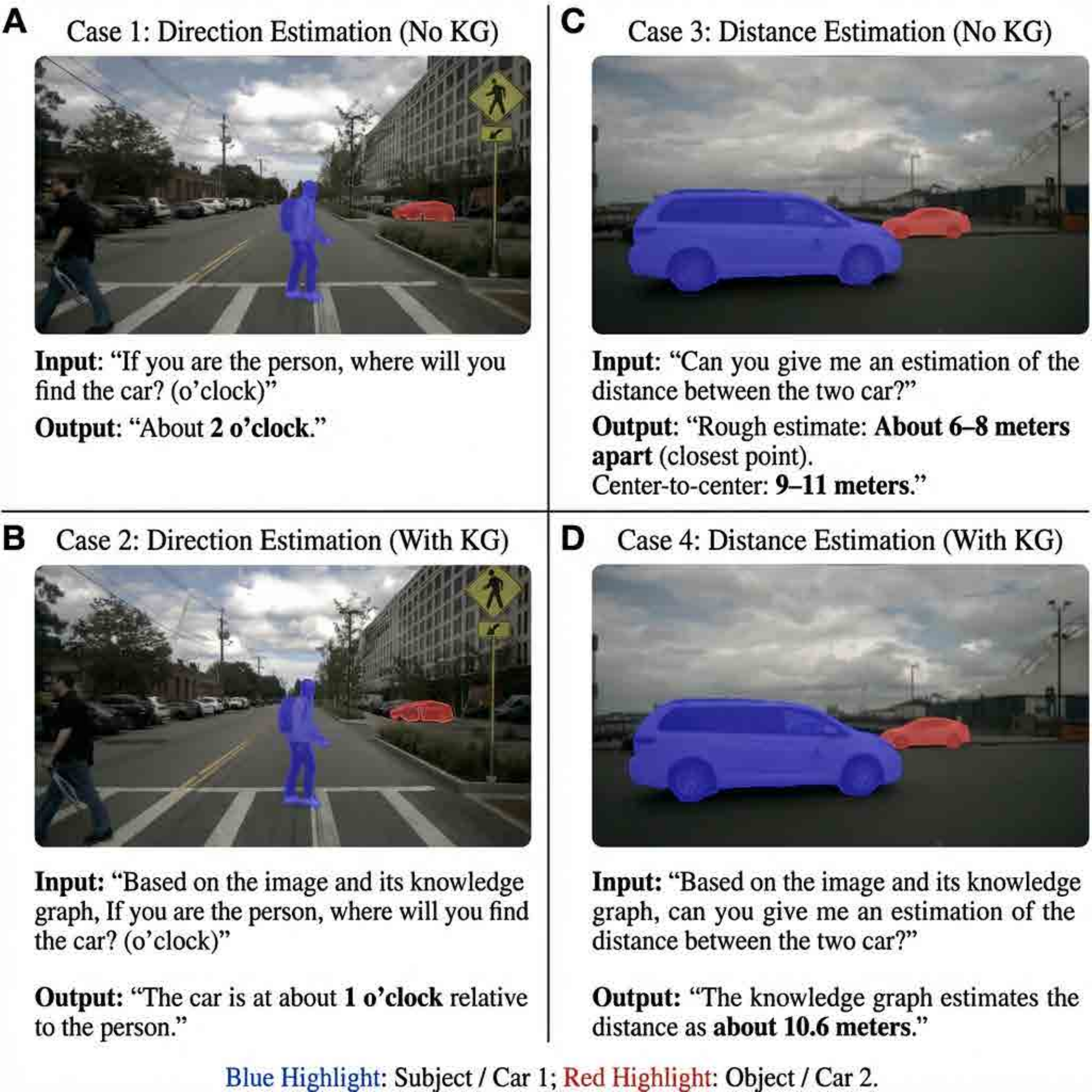}
  \caption{Qualitative examples of spatial QA based on street view imagery: comparison of GPT-5.4 and DM-KG enhanced reasoning results}
  \label{fig:qa}
\end{figure}

To objectively evaluate the spatial reasoning performance of the proposed method, we compared it against GPT-5.4 and SpatialRGPT. SpatialRGPT is explicitly designed for spatial visual reasoning. We use the public spatial cognitive QA benchmark SpatialRGPT-Bench, with comparative experiments conducted across more than 200 QA images for direction judgment and distance estimation. Our input queries remained consistent, and input images consisted of street view imagery annotated with entity masks. As illustrated in Figure~\ref{fig:qa}, qualitative observations during QA testing on GPT-5.4 indicate that, utilizing the external structured DM-KG as prior input significantly enhances the VLM's answer accuracy when confronting complex spatial queries.

Table~\ref{tab:comparison} details the quantitative comparison results between our method and baseline models on the distance estimation and direction judgment subtasks within the SpatialRGPT-Bench dataset. In the distance estimation task, both our method and GPT-5.4 achieved a 56.3\% success rate. However, the Mean Absolute Error (MAE) of our method was significantly reduced from 5.60~m to 3.86~m, with a relative reduction of 31.1\%. The Absolute Relative Error (AbsRel) also decreased from 0.373 to 0.309 with a 17.2\% reduction. This reveals that while both models share similar success rates in qualitative near-far discrimination, our method effectively corrects systematic biases introduced when VLMs rely on semantic category prior sizes for perspective estimation. Notably, the integration of explicit depth information also dramatically improves the numerical precision of absolute distance estimation when processing near-to-mid distance objects.

\begin{table}[htbp]
  \caption{Performance comparison of distance estimation and direction judgment tasks based on SpatialRGPT-Bench}
  \label{tab:comparison}
  \centering
  \resizebox{\columnwidth}{!}{
  \begin{tabular}{lcccccc}
    \toprule
    & \multicolumn{3}{c}{Distance} & \multicolumn{3}{c}{Direction} \\
    \cmidrule(lr){2-4} \cmidrule(lr){5-7}
    & Success & MAE & AbsRel & Success & MAE & AbsRel \\
    & Rate(\%) & (m) & & Rate(\%) & (deg) & \\
    \midrule
    SpatialRGPT & 46.7 & 4.52 & 0.330 & \textbf{96.7} & 7.71 & 0.256 \\
    GPT-5.4      & 56.3 & 5.60 & 0.373 & 90.7 & 9.56 & 0.319 \\
    DM-KG       & \textbf{56.3} & \textbf{3.86} & \textbf{0.309} & 95.3 & \textbf{3.27} & \textbf{0.109} \\
    \bottomrule
  \end{tabular}}
\end{table}

In the direction judgment task, our method achieved a 95.3\% success rate, slightly below the 96.7\% achieved by SpatialRGPT, but demonstrated a substantial advantage in error reduction. MAE plummeted from 7.71 degrees to 3.27 degrees with a drop of 65.8\%, and AbsRel sharply fell from 0.256 to 0.109 with a 65.8\% drop. These data demonstrate that encoding and metricizing the directional relationships between entities using DM-KG provides the VLM with highly precise geometric reference frames, thereby refining direction judgments. Even without specific fine-tuning of the VLM for visual spatial features, the introduction of external spatial priors grants it geometric calculation precision rivaling that of dedicated spatial VLMs.

As Table~\ref{tab:comparison} indicates, our method exhibits distinct performance characteristics in spatial QA compared to the purpose-built SpatialRGPT. The fundamental reason for this advantage lies in error accumulation within SpatialRGPT. Its internal reasoning is inevitably affected by camera focal length estimation errors, which compound with depth estimation errors to cause an overall deviation in point cloud projection. In contrast, our method provides a precise geometric baseline for directional relationships by explicitly injecting camera focal length parameters and performing analytical calculations via the pinhole camera model. The slightly lower direction judgment success rate (95.3\%) compared to SpatialRGPT (96.7\%), may be attributed to minor centroid calculation errors of panoptic segmentation masks or slight deviations in median depth extraction under extreme occlusion scenarios.

\section{Conclusion}
\label{sec:conclusion}

In this study, we introduce DM-KG, an explicit, structurally grounded spatial knowledge graph framework for enhancing VLM spatial reasoning in street view imagery. By integrating systematic monocular depth model selection with joint panoptic segmentation, DM-KG overcomes the limitations of previous models that rely on implicit spatial learning. Implicit end-to-end approaches often fall short in high-stakes spatial tasks due to their opaque reasoning and untraceable errors. Our framework addresses this by explicitly recovering entity-level 3D coordinates and encoding clock azimuths and Euclidean distances into a knowledge graph. By injecting this graph as a geometric prior during VLM prompting, we shift the model away from statistical guessing toward logical deduction based on explicit geometric constraints, providing a robust and interpretable solution for open-scene geographic VQA.

However, the current method has two core limitations. First, extracting spatial relationships based solely on point-to-point associations is restrictive. Representing linear or planar targets with a single median depth point fails to comprehensively describe its geometric characteristics, potentially introducing systematic biases. Second, depth accuracy relies heavily on precise focal length calibration; missing or erroneous image metadata significantly restricts the performance of our framework. Future research will focus on depth representation for linear and planar entities and adaptive focal length calibration in parameter-free scenarios, aiming to further enhance the framework's robustness and generalizability in complex real-world environments.


\clearpage
\footnotesize
\section*{Supplementary Material}

\textbf{1. Pixel-level Heatmap Overlay Visualization of MAE(m) and AbsRel across Models} (In each grid, top-left is Metric3Dv2, top-right is DepthAnythingV2 (no f), bottom-left is DepthPro, and bottom-right is UniDepthV2; red indicates larger errors.)

\begin{figure}[htbp]
  \centering
  \includegraphics[width=0.95\columnwidth]{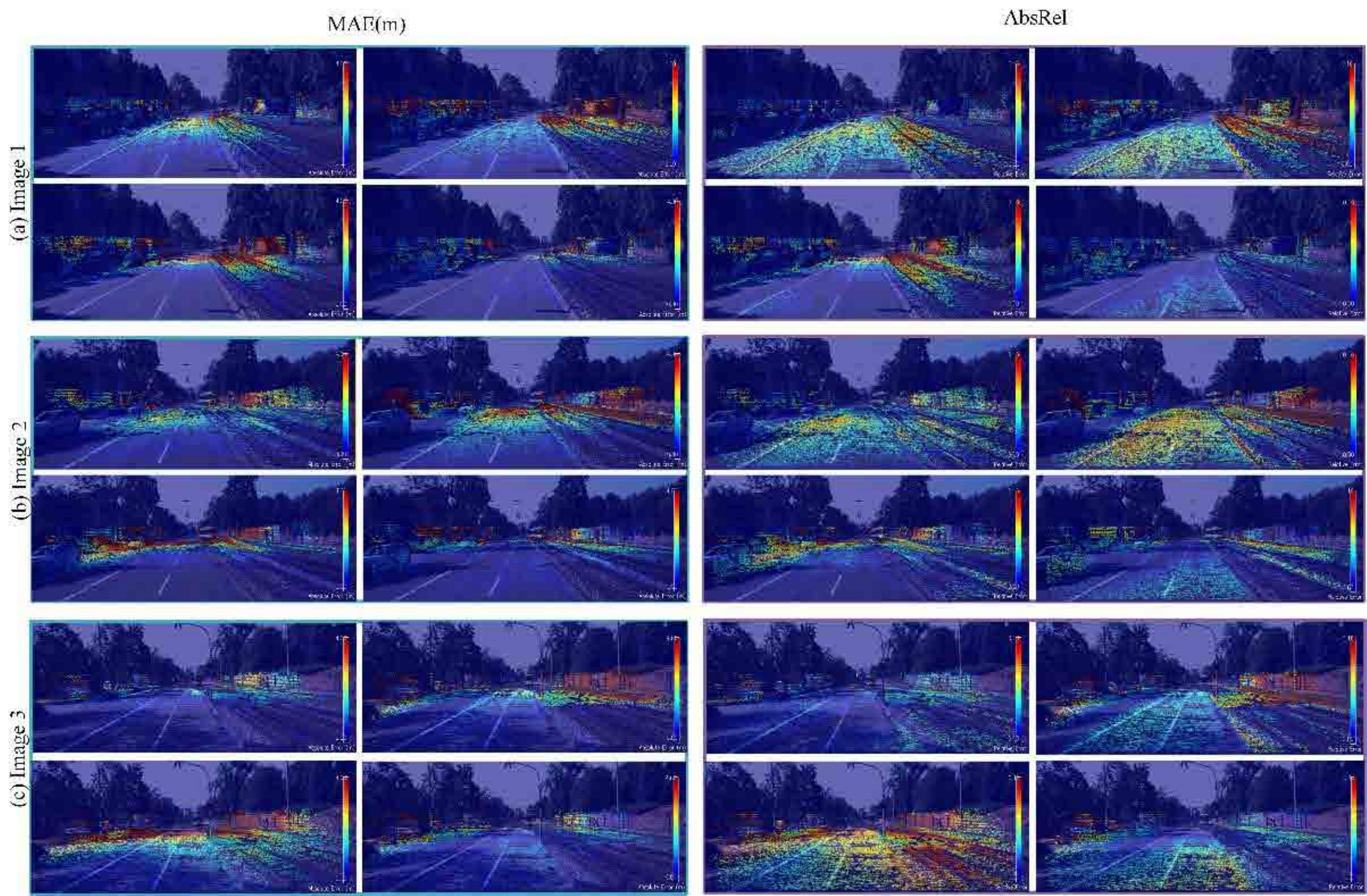}
\end{figure}

\textbf{2. Visualization of Entity-Level 3D Coordinate Extraction Pipeline}

\begin{figure}[htbp]
  \centering
  \includegraphics[width=0.85\columnwidth]{image11.pdf}
\end{figure}

\end{document}